\pdfoutput=1
\documentclass[11pt]{article}

\usepackage[final]{mtsummit25}
\usepackage{copyright}
\usepackage{times}
\usepackage{latexsym}

\usepackage{caption}
\usepackage{afterpage}
\usepackage{placeins}
\usepackage{float}
\usepackage{graphicx} 
\usepackage{multicol}
\usepackage{dblfloatfix}

\usepackage[T1]{fontenc}
\usepackage[utf8]{inputenc}
\usepackage{CJKutf8}
\usepackage{microtype}

\usepackage{inconsolata}

\usepackage{graphicx}
\usepackage{amsmath}
\usepackage{booktabs}

\usepackage{todonotes}
\title{Optimising ChatGPT for creativity in literary translation:\\A case study from English into Dutch, Chinese, Catalan and Spanish}

\author{
  \textbf{Shuxiang Du\textsuperscript{1}\addtocounter{footnote}{1}\thanks{Equal contribution}}, 
  \textbf{Ana Guerberof Arenas\textsuperscript{1}\textsuperscript{$\dagger$}}, 
  \textbf{Antonio Toral\textsuperscript{2}\textsuperscript{$\dagger$}} \\
  \textbf{Kyo Gerrits\textsuperscript{1}}, 
  \textbf{Josep Marco Borillo\textsuperscript{3}\addtocounter{footnote}{1}} \\
  \textsuperscript{1}Centre for Language and Cognition, University of Groningen \\
  \textsuperscript{2}Departament de Llenguatges i Sistemes Informàtics, Universitat d'Alacant \\
  \textsuperscript{3}Departament de Traducció i Comunicació, Universitat Jaume I \\
  \texttt{sophie321du@gmail.com}
}

\begin{document}
\maketitle
\begin{abstract}
This study examines the variability of ChatGPT's machine translation (MT) outputs across six different configurations in four languages, with a focus on creativity in a literary text. We evaluate GPT translations in different text granularity levels, temperature settings and prompting strategies with a Creativity Score formula. We found that prompting ChatGPT with a minimal instruction yields the best creative translations, with "Translate the following text into [TG] creatively" at the temperature of 1.0 outperforming other configurations and DeepL in Spanish, Dutch, and Chinese. Nonetheless, ChatGPT consistently underperforms compared to human translation (HT). All the code and data are available at \texttt{https://github.com/INCREC/Optimising}. 
\end{abstract}

\section{Introduction}

The intersection of artificial intelligence (AI) and creativity in the domain of translation presents a fascinating and challenging field for research. Even if the development of machine translation (MT) technologies, especially through the advent of Large Language Models (LLMs) like ChatGPT, has reshaped the landscape of the language industries, there remains a notable gap in the creative capacities of MT outputs in comparison to that of professionals~\citep{karpinska-2023-new}. This type of translation, often applied to literary texts, requires not just the accurate conveyance of meaning but also the preservation of style, tone, and creative nuances inherent in the source text to create an effect on the reader that is not purely information driven. Since new models offer a dialogic capacity, we explore in this paper the best set of variables to generate the most creative translations using ChatGPT.

We investigate various configurations of ChatGPT, including different text granularities, temperature settings, and prompting strategies, alongside a comparison with %NMT translations (DeepL) 
translations by neural machine translation (NMT) systems, and human translations as references. The primary aim is to evaluate how these configurations impact the creativity and quality of the translations produced, using both a manual creativity scoring system and automatic evaluation metrics. 
The experiment involves translating a short science fiction story by Kurt Vonnegut, "2BR02B", from EN (English) to ZH (Chinese), NL (Dutch), CA (Catalan), and ES (Spanish). The translations are manually annotated to assess creative shifts (CSs) and errors, providing a detailed analysis of how ChatGPT in different configurations handles the nuanced demands of literary translation. Thus, central to this research are the questions: 

RQ1: What is the variability in MT outputs from ChatGPT under different settings?

RQ2: What is the optimal prompting setting for the most creative MT output using ChatGPT? 

\section{Related Work}\label{s:rel_work}

LLMs like ChatGPT have demonstrated promising performances in natural language processing tasks \citep{kalyan-2023}. LLMs such as IOL-Research, Unbabel Tower 70B, and Claude-3.5-Sonnet are the top performing MT systems submitted to the last edition of WMT's general translation task \citep{kocmi-etal-2024-findings}. ChatGPT for MT has demonstrated promising applications to help users translate specific contents or entire documents, especially between high-resourced languages (\citealp{jiao-2023-new}, \citealp{hendy-2023-new}). However, whether it outperforms NMT systems or commercial MT systems is still under debate \citep{kalyan-2023}.

The research community has investigated the effectiveness of ChatGPT for MT in different aspects. \citet{gao-2024-new} focused on developing advanced prompting strategies by including additional information like task, domain, and syntactic information like PoS (parts of speech) tags. The researchers tested the language pairs English \(\leftrightarrow\) Spanish, English \(\leftrightarrow\) French, and Spanish \(\leftrightarrow\) French in the domains of news, e-commerce, social, and conversational in a sentence level. They concluded that including appropriate information about the input text in the prompt, such as specifying translation task or context domain, can improve the performance of ChatGPT. ChatGPT has a higher BLEU \citep{10.3115/1073083.1073135} score in four out of the six language pairs when compared to Google Translate (GT) and DeepL Translate (DeepL) with their proposed advanced prompting strategies. 

In terms of text granularity levels, \citet{wang-2023-new} examined the performances of ChatGPT for document-level translation, covering three language pairs (Chinese \(\Rightarrow\) English, English \(\Rightarrow\) German, and English \(\Rightarrow\) Russian) in seven domains (news, social, fiction, Q\&A from an online forum, TED, Europarl, and subtitle). The researchers reported that ChatGPT does well when the sentences in the document are combined and given at once to the model. With this prompting strategy, it exhibited better performances than commercial MT systems according to human evaluation and also outperformed most document-level NMT methods in terms of d-BLEU scores. 

Temperature is a hyperparameter in %large language models 
LLMs that regulates the randomness in text generation by adjusting the probability distribution of potential next words \citep{peeperkorn-2024-new}. Decoding with higher temperatures displays greater linguistic variety, while low values tend to generate grammatically correct and more deterministic text \citep{ippolito-2019-new}. \citet{peng-2023} explored the impact of temperature, task, and domain information on the translation performance of ChatGPT. In the translation of English, Chinese, German, and Romanian of biomedicine, news, and e-commerce texts, the study showed that ChatGPT performance degraded with an increase in temperature in terms of both BLEU and COMET 
\citep{rei2020comet-new}
scores, and hence it was recommended to use a lower temperature (recommended is 0 for their test set). Additionally, including task and domain information in the prompt enhanced the translation performance of ChatGPT consistently for both high- and low-resource languages in their research. 

As MT technology advances, there is growing interest in exploring how well these systems can handle the complexities of literary translation. With respect to LLMs, %GPT translation, 
\citet{karpinska-2023-new} evaluated the performance of ChatGPT in translating literary paragraphs across 18 linguistically diverse language pairs. The authors experimented with three different prompting strategies, namely translating sentence by sentence in isolation, translating sentence by sentence in the presence of the rest of the paragraph, and translating the entire paragraph at once. According to human evaluation, when translating entire paragraphs, ChatGPT produced translations of significantly higher quality compared to other strategies and commercial systems. However, critical errors such as content omissions still occur. The findings suggest that while ChatGPT can leverage larger context units like paragraphs to enhance translation quality, this is yet not sufficient on their own for high-stakes applications like literary translation where nuanced understanding and stylistic consistency are crucial.

The challenges of literary MT lie not only in the performance of the systems but also in the evaluation of the results. \citet{fonteyne-2020-new} provided an in-depth evaluation of the quality of a novel translated by %neural machine translation (NMT) 
NMT
from English \(\Rightarrow\) Dutch. Unlike traditional sentence-level evaluations, this study emphasized the importance of document-level analysis to better assess the coherence and cohesion of translated texts, which are crucial in literary translations. It utilized an adapted version of the SCATE error taxonomy \citep{tezcan-2017-new}, which considers errors at both the sentence and document levels. Again, the findings suggested that while NMT can produce a substantial portion of error-free translations, significant errors remain, particularly with complex elements like style and coherence that are vital to literary texts. Therefore, it is important to consider metrics other than error annotation when evaluating literary texts. 

In the studies of ChatGPT for translation, most evaluations focus on automatic metrics like COMET and BLEU, while the specific aspect of creativity has hardly been touched upon. This could be because creativity in translation is hard to measure. \citet{bayer-2009-new} proposed a framework for assessing translational creativity based on the concepts of novelty, acceptability, flexibility, and fluency. Novelty in translation is characterized by three main aspects: exceptional performance that significantly surpasses routine translation activities, uniqueness or rarity within a specific corpus of translations, and non-obligatory translational shifts that indicate a high level of translator engagement and creativity. Acceptability is defined as “skopos adequacy” (\citealp[2]{bayer-2009-new}). This emphasizes that a creative translation must not only be innovative but also appropriate and useful within the context for which it is intended. These novelty and acceptability aspects of the framework are largely adopted in this research.

\citet{bayer-hohenwarter-2011} defines creative shifts as transformative operations in translation that deviate from the direct replication of the source text. These shifts are categorized into three types: abstraction, where translators generalize specific details from the source; modification, which involves alterations to better suit the cultural or contextual needs of the target text; and concretization, where translators add specific details not explicitly mentioned in the source text. Bayer-Hohenwarter proposed a systematic methodology to measure creativity in translation by identifying and analyzing these creative shifts. She defined specific “units of analysis” within the texts, identifying both “creativity units” (requiring high problem-solving capacity) and “routine units” (relatively straightforward translation tasks). The results were quantified by calculating the proportion of creative shifts versus literal reproductions. The study also examined the relationship between the frequency of creative shifts and the overall quality (acceptability) of the translations. There was a general trend suggesting that translators who produced more creative shifts also produced higher-quality translations. However, this was not a strict correlation, as some creative shifts led to errors, particularly among less experienced translators.

%\citet{bayer-hohenwarter-2020} reviewed the role of creativity in translation, exploring cognitive models and theoretical frameworks such as Skopos Theory, visualization, and switch competence. Additionally, they analyzed a case study from an online translator forum to illustrate how translators approach creativity in practice, highlighting the tensions between novelty and adequacy in translation choices. The study introduces the concept of "Facebook brainstorming" as a model of distributed translation work and stresses the translator's responsibility in selecting final solutions. They underscore the increasing reliance on digital tools, positioning creative adaptation as a key differentiator for human translators. 

\citet{jbp:/content/journals/10.1075/ts.20035.gue, jbp:/content/journals/10.1075/ts.21025.gue} created a formula (see section 3) to quantify creativity in translations, offering a measurable way to assess and compare the creative output of different translation modalities, including MT. Their study involved the translation of literary texts from English to Catalan and Dutch. The texts were translated by professionals, post-edited by professionals, and machine translated. By applying a creativity score to the translations, they found MT outputs to be less creative than professional translations and that they limited the translator’s creativity in post-editing. The quantification framework they established is used in this research.

In their Master thesis \citet{du2024thesis} use this creativity index in an evaluation of ChatGPT translations of a literary text in the %language pair of 
English \(\Rightarrow\) Chinese
translation direction.
They investigated different set-ups of ChatGPT including levels of text granularities, different temperatures, prompting strategies, and few-shot prompting. The findings indicated that the quality and creativity of ChatGPT translations vary across these configurations. The best setting in their study was a document-level translation with a temperature of 1.0 and a direct prompt to be more creative. In this paper, we replicate the experiment with more languages and a more in-depth analysis.

\section{Methodology}
In this section, we explain the %content
source text (ST) used, how the target texts (TTs) were generated in the different phases of our experimentation, as well as the data annotation and analysis process. 

\subsection{Source Text}

The study utilized a curated dataset comprising different translations of a short science fiction story by Kurt Vonnegut: \textit{2BR02B}\footnote{\url{https://www.gutenberg.org/ebooks/21279}}~\citep{vonnegut-1999}. 
The story is a short science fiction piece set in a future society where aging has been cured and the population is strictly controlled to remain at forty million. Individuals must volunteer for death to allow new births. It revolves around a family about to give birth to three kids and therefore in need of three volunteers to die.

This story was selected for three reasons: A) we have an existing corpus of annotations on the units of creative potential in the story \cite{jbp:/content/journals/10.1075/ts.21025.gue}, B) to our knowledge it has not been translated into the target languages to date\footnote{Not found in the Unesco Translationum database \url{https://www.unesco.org/xtrans/bsform.aspx} nor on National Library of China \url{https://www.nlc.cn/web/index.shtml} } and therefore we assume it has not been used in the training data of ChatGPT, and C) it requires a high level of translation creativity. 

The story was processed in Python to be broken into separate paragraphs. The text overall contains 123 paragraphs, 234 segments and 2548 words. There are 185 units of creative potential (UCP) in total, annotated by two experienced translators and researchers in the previous study \cite{jbp:/content/journals/10.1075/ts.21025.gue}. These are units in the ST that are expected to require translators to use problem-solving skills, as opposed to those that are regarded as routine units with little creative potential (\citealp{bayer-hohenwarter-2011}). 

%\The units are classified as follows: A) metaphors and original images, B) comparisons, C) idiomatic phrases, D) wordplay and puns, E) onomatopoeias, F) colloquial language (cursing, slang, for example), G) phrasal verbs, H) cultural and historical references, I) neologisms, J) lexical variety (number of adjectives before the noun or use of adverbs, for example) K) expressions specific to linguistic variant (for example, American English or British English) L) unusual punctuation, M) rhyme and metrics, N) proper names, and O) treatment (formal, informal) (ANONYMIZED, 2022, 8-9).

\subsection{Target Text}

For the target text (TT), we used the model gpt-4o-2024-08-06 with the ChatGPT API\footnote{Version 1.54.4, i.e. the latest when we started out experiments.} to translate the text into ZH, NL, ES and CA. This version is chosen for three reasons: A) %it is to date the latest stable model of ChatGPT and represents 
it was the latest stable model of ChatGPT when we started our experimentation, thus representing state-of-the-art performance, B) according to OpenAI\footnote{\url{https://openai.com/index/hello-gpt-4o/}} , this version performs better on text in non-English languages, C) in terms of data training, the cost of this version is relatively lower and the speed is faster when compared to ChatGPT-4. 

Due to limited capacity and the exploratory nature of this experiment, we decided to annotate a subset of the text. We selected a series of UCPs that were previously singled out by two annotators in the ST to ensure a better representation of the creative potential of this text. In the end, 54 UCPs, present in 48 separate sentences with a total of 602 words were selected for the annotation task in the TT. To prepare the sentence-aligned files, we manually post-processed the text by extracting the 48 sentences in each translation.
 
Each translation of the sentences in the TT was manually annotated for a detailed comparative analysis of creativity across different translations. The annotators were four of the researchers that are experienced translators or have a language related Master degree in the selected language combinations: there was therefore one annotator per language combination.

As the baseline of the study, DeepL %\todo{motivate choice?}
has been chosen to compare with ChatGPT. The reason is that in the preliminary experiment \citep{du2024thesis} it offered a more pleasant-to-read translation than other NMT systems like Google Translate. Since DeepL is not available for CA, we used two popular NMT systems for this target language: Softcatalà's \textit{Traductor}\footnote{\url{https://www.softcatala.org/traductor/}} and Google Translate.\footnote{\url{https://translate.google.com/}}
%Because it is not available in Catalan, we used Softcatalà and Google Translate for ENCA.

%\todo{Something is missing here}

\subsection{Data Collection}

\begin{figure}[htbp]
    \centering
    \includegraphics[width=0.45\textwidth]{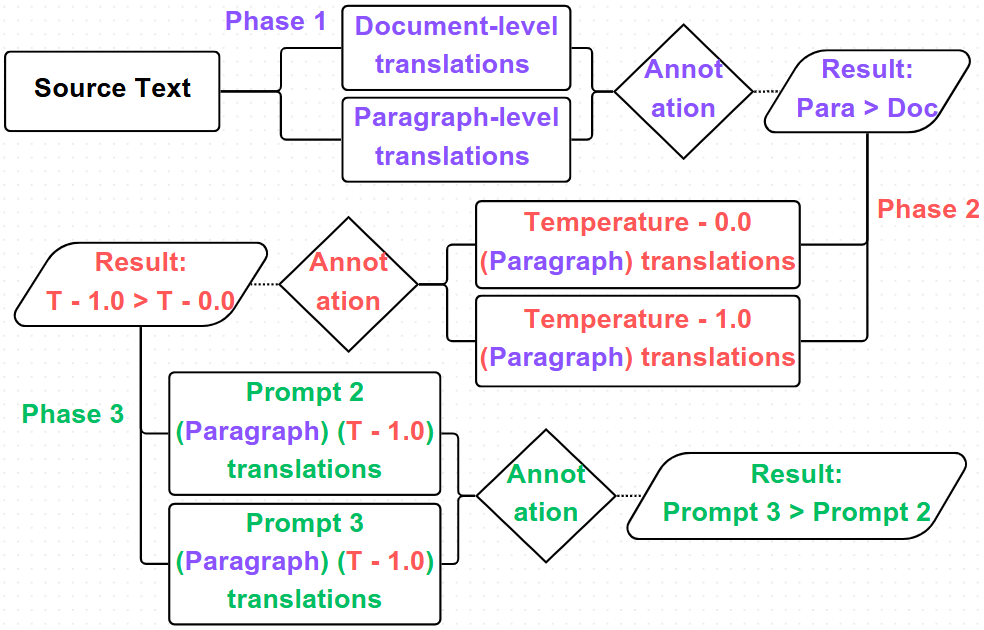}
    \caption{Workflow for ZH and NL}
    \label{fig:1}
\end{figure}

In this experiment, we try a range of text granularities, temperature settings, and zero-shot prompting strategies based on \citet{du2024thesis} master project to generate translations %from 
with ChatGPT. The experiments and annotations were conducted between October 2024 and January 2025. Figure \ref{fig:1} shows an overview of the NL and ZH workflow process as an example. The workflow in each phase slightly differs for the other two languages (CA and ES).

\subsubsection{Phase 1. Text Granularity}

The variable in the first phase is text granularity. We translated the text at both %sentence level and paragraph level.
paragraph level (setting 1a) and document level (1b).
At the paragraph level, we entered the same prompt for each paragraph in the story, with each request done separately %opened in a new chat 
to avoid context interference. At the document level, we entered the same prompt followed by the entire story in one request. It is worth noting that 14 of the 48 sentences involved in the evaluation process were single-sentence paragraphs.%\todo{How many from the ones in the experiment?} 

The prompts used in phase 1 are: 

\textbf{Prompt 1} (1a): "Translate into [TG]: [Input]"

\textbf{Prompt 1} (1b): “Translate the following text into [TG]: [Input]”

On the one hand, having the whole document available offers more context, which should be useful for its translation. On the other, recent research has shown that translation performance decreases with the length of the input text~\citep{peng2024investigating}.

After the evaluation of the creativity score for these two outputs (see Section~\ref{s:results_granularity}), we proceeded with the better granularity method for each target language to further experiment with different temperatures and prompts. %For example, 
Namely, in CA and ES, the document-level translations were assessed to be better, thus the following experiment in CA and ES were conducted at document level (1b), while the opposite was the case for ZH and NL (1a).

\subsubsection{Phase 2. Temperature}

The temperatures selected are 0.0 (2a) and 1.0 (2b). In the API setup, ChatGPT's temperature can range from 0.0 to 2.0. The default temperature setting of ChatGPT is officially stated to be set to 1.0\footnote{\url{  https://platform.openai.com/docs/api-reference/chat/create}}.
However, according to our experience the default value seems to be 0.0. Namely, we used the automatic evaluation metric chrF \citep{popovic-2015-chrf} to examine how similar the translations produced using different temperature values are to the translations produced in phase 1, which used the default setting for temperature. For NL, for example, the chrF result decreased as the temperature went up: 88 (temperature=0.0), 86 (0.5), 83 (1.0), 82 (1.1) and 81 (1.2). This means that ChatGPT is not deterministic, even with temperature 0.\footnote{If it was deterministic, then chrF's score when comparing phase 1's translation with phase 2's translation with temperature=0 would have been 100,} We then chose the value 0 in phase 2 to see the effect of non-determinism. 

On the other hand, the higher the temperature is, the more creative the text is expected to be. However, the highest value we chose was not the maximum offered by the API (2.0) but 1.0.
This is because we noticed that at higher temperature values, there are more instances of %bugs and 
“word vomit” in the output which makes the text incoherent and impossible to read.\footnote{We tried three values of temperature higher than 1.0 (1.1, 1.2 and 1.5) and noticed severe issues with values 1.1 (CA), 1.2 (ES), 1.5 (NL and ZH). We speculate that the reason why ChatGPT has issues in CA and ES at a lower temperature values than NL and ZH is because for the former the document is translated at once.
}
Therefore, at temperature 1.0 the system is most likely to generate more creative content while not suffering from word vomit. 

The same prompt as in phase 1 was used and we proceeded with the best temperature setting after evaluation (see Section~\ref{s:res_temperature}). For ES, NL, and ZH, we proceeded with temperature 1.0 (2b), while for CA we proceeded with 0.0 (2a).

\subsubsection{Phase 3. Prompting Strategies}

The zero-shot prompting strategies we designed %for ChatGPT
included prompting with the specific domain information, i.e. author and genre (3a), and prompting with direct instructions to generate creative outputs (3b). The final prompts are as follows:

\begin{itemize}
    \item \textbf{Prompt 2} (more info about genre and author, 3a): Translate the following text into [TG] taking into
consideration that this is (from) a science fiction story by Kurt Vonnegut: [input]
    \item \textbf{Prompt 3}  (request of creativity, 3b): Translate the following text into [TG] creatively: [input]
\end{itemize}

\subsection{Data Annotation}

Following data collection, each sentence of each translation was manually annotated in terms of acceptability and novelty, as discussed in section~\ref{s:rel_work}. The annotators were blind to the specific setting they were evaluating.

Acceptability was measured according to the number and severity of errors in the TTs based on the harmonized DQF-MQM Framework \citep{lommel-2014}. %\todo{Reference to DQF-MQM}. 
The severity of each error was marked as: Neutral (0 points for repeated errors or preferences), Minor (1 point), Major (5 points), and Critical (15 points). Minor refers to errors that do not lead to loss of meaning and do not confuse or mislead the reader but are noticeable, hence they decrease stylistic quality, fluency or clarity, or make the content less appealing. Major refers to errors that may confuse or mislead the reader or hinder the understanding of the text due to significant change in meaning or because errors appear in a visible or important part of the content. Critical refers to errors that may misrepresent or damage the reputation of the author or publishing house, causes the text to stop working as a literary artefact and affect the communicative flow, or if the language is perceived as offensive (when unintended), but also, if the text departs from the source text in such a significant way that has a large impact on the understanding of the entire story. 

For example: the title of the story, "2BR02B", is a play on words on the famous quote \textit{To be or not to be} in Shakespeare\textquotesingle s \textit{Hamlet}. If left in English in ES and CA, the understanding of the entire story is compromised, and it is therefore considered a Critical error. If the word \textit{business} is translated literally in a context where it does not refer to a commercial activity but to a person's concern, then this can be considered a Major error because the entire text is understood albeit with certain difficulties. Finally, a spelling mistake would be considered Minor.

For novelty, the 
translation
solutions to the UCPs selected were annotated in the TTs. All translations that deviate from the ST that are neither the exact reproduction of the ST nor an omission nor an error count as CS and are classified in the following manner: Abstraction refers to instances when translators use more vague, general or abstract solutions. Concretization refers to instances when the TT evokes a more explicit, more detailed, and more precise idea or image. Modification refers to instances when translators use a different solution in the TT (e.g. express a different metaphor without the image becoming more abstract or concrete).

For example: if the title of the story, \textit{2BR02B}, is translated into CA as \textit{C-O-N-O-C}, a play on words that evocates \textit{Ser o no ser}, the standard Catalan phrase, this would be considered a Modification and 
classified as CSM. While in NL, the title \textit{2BR02B} is left as is, as the standard phrase in Dutch %is
remains
\textit{To be or not to be}. This is then considered a Reproduction, and classified as such. As this exemplifies, Reproductions are not errors by default, although some UCPs that are not translated might be considered as containing an error.

In the process of annotation, the translation of the 54 UCPs was assessed. For each translated UCP, the annotator decided if the resulting TT was a CS, an omission (O), a reproduction (R), or if it was impossible to classify (E). The CSs were further classified into abstraction (CSA), concretization (CSC), and modification (CSM). Each of the 48 sentences were annotated for errors according to the severity criteria described. The total number of CSs and Error points was used for the creativity index, introduced next.

\subsection{Data Evaluation}\label{s:data_eval}
Acceptability and novelty are combined into a single score using the creativity index (CI) formula:
\small
\begin{equation}
\text{CI} = \\
\left(
\frac{\#CSs}{\#UCPs} 
- 
\frac{\#\text{error points}}{\#\text{words in ST}}
\right) \notag \\
\times 100
\end{equation}
\normalsize

The index considers both novelty (CSs) and appropriateness or acceptability (errors), enabling a quantifiable comparison between different translation modalities \cite{jbp:/content/journals/10.1075/ts.20035.gue, jbp:/content/journals/10.1075/ts.21025.gue}. 

Apart from the creativity index, 
we used a number of automatic evaluation metrics (AEMs): BLEU, chrF, TER \citep{snover-etal-2006-study}, COMET and COMET-Kiwi \citep{rei2022cometkiwi-new}. %\todo{Antonio. add citations Sophie. I have added for kiwi and TER. The other three are in earlier text}
The first three are string-based\footnote{We compute them with sacrebleu 2.5.1} while the last two are based on multilingual language models.\footnote{We used models wmt22-comet-da and wmt22-cometkiwi-da, respectively.}
Another distinction is that the first four evaluate a translation with respect to a reference translation (see Section~\ref{s:ht}),\footnote{COMET takes into account also the ST.} while the last one does so with respect to the source text.
Since we do not have a reference translation for ZH, only COMET-Kiwi was used.

%\textcolor{blue}{
%These AEMs work at sentence level, and therefore they require that all the files given follow a sentence-by-sentence correspondence. To fulfil this some manual post-processing was required. This involved merging a few paragraphs for CA and ES and splitting a segment into two for NL.\todo{This paragraph can be deleted if you think this is too detailed}
%}
%COMET-QE was used to compare the MT outputs. COMET computes scores by leveraging pre-trained transformer models to encode both the source text and the machine-translated text into high-dimensional embeddings \citep{rei-2020}. This score, ranging from 0 to 1 from the lowest quality to the highest quality, reflects the predicted quality of the translation in terms of fluency, adequacy, and fidelity to the source text, without requiring a reference translation. This would shed light on the difference in the general translation outputs other than errors or solutions of UCPs. 
%\todo{Antonio. Add BLEU, chrf, TER %and the manual post-processing to get sentence-aligned text
%}

\subsection{Human Reference}\label{s:ht}
%\todo{Antonio. Move this earlier? I think it ideally position is 3.3. In any case it should go before 3.5} \todo{Sophie. But in table 1 there are mentions of Creativity Index so maybe it should go after 3.5?}
Since this experiment utilizes a dataset from the \citet{jbp:/content/journals/10.1075/ts.21025.gue} project, we had access to translations created by professionals in EN$\Rightarrow$CA, EN$\Rightarrow$NL and EN$\Rightarrow$S,\footnote{The Spanish translation was not analyzed in the previous project, but was translated by the researcher to be used as a reference. This version was then annotated for errors by a professional literary translator.
} but unfortunately not for EN$\Rightarrow$ZH. Table \ref{tab:table1} shows the scores for the selected UCPs for these languages.

\begin{table}[htbp]
\centering
\begin{tabular}{@{}ccccc@{}}
\toprule
     & \# CSs & \# Errors & Error Points & CI  \\ \midrule
ENCA & 21      & 2        & 2          & 40 \\
ENES & 22      & 6        & 6          & 40 \\
ENNL & 29      & 17        & 25          & 50  \\ \bottomrule
\end{tabular}
\caption{Creativity Index in Human Reference}
\label{tab:table1}
\end{table}

The results for ENCA and ENES were annotated by a professional literary translator, while the ENNL was annotated by a different one for that language pair, and this could account for the differences in judgement, although, of course, this could also mean that there are differences in the quality provided by the translators. One aspect to note here is that while annotating the UCPs, the reviewers also remarked that the entire segments contained other CSs. For example, in ES and CA, the translators changed the name of the characters to be able to create meaningful play on words that were present in the ST.

\section{Results}

The following subsections contain the results obtained in each of the phases explained in the methodology.
%Table 1, Table 2, Table 3, and Table 4 provide an overview of the creativity scores.
Detailed annotations per language are provided in %For more details in the annotations, see 
Appendix \ref{a:human_annotations}.

\subsection{Phase 1. Text Granularity}\label{s:results_granularity}
Table \ref{tab:table3} shows the results for phase 1, ChatGPT outputs at paragraph (1a) and document level (1b) were compared.

\begin{table*}[htbp]
    \small
    \centering
    \begin{tabular}{@{}ccccccccc@{}}
    \toprule
        & \multicolumn{4}{c}{Paragraph (1a)}                              & \multicolumn{4}{c}{Document (1b)}                               \\ \midrule
        & \# CSs      & \# Errors   & Error Points & Score           & \# CSs      & \# Errors   & Error Points & Score           \\ \midrule
    %ENCA & 5           & 51          & 199          & -23.80          & \textbf{4}  & \textbf{51} & \textbf{158} & \textbf{-18.84} \\
    ENCA & \textbf{5}           & \textbf{51}          & 199          & -23.80          & \textbf{5}  & \textbf{51} & \textbf{158} & \textbf{-16.99} \\
    ENES & 6           & 59          & 253          & -31.00          & \textbf{10} & \textbf{52} & \textbf{219} & \textbf{-18.00} \\
    ENNL & \textbf{15} & \textbf{61} & \textbf{108} & \textbf{9.84}   & 12          & 68          & 174          & -6.68           \\
    ENZH & \textbf{11} & \textbf{51} & \textbf{187} & \textbf{-10.69} & 10          & 56          & 298          & -30.98          \\ \bottomrule
    \end{tabular}
    \captionsetup{justification=centering, font=small}
    \caption{Creativity score for two different text granularities: paragraph and document. \\The best score per language and criterion is shown in bold.}
    \label{tab:table3}
\end{table*}

%\begin{table}[htbp]
%\centering
%\begin{tabular}{@{}ccccc@{}}
%\toprule
%     LP & \# CSs      & \# Errors   & Error Points & Score \\
%     \hline
%     & \multicolumn{4}{c}{Paragraph} \\          
%     \hline
%ENCA & 5           & 51          & 199          & -23.80          & \textbf{4}  & \textbf{51} & \textbf{158} & \textbf{-18.84} \\
%ENCA & \textbf{5}           & \textbf{51}          & 199          & -23.80          \\
%ENES & 6           & 59          & 253          & -31.00          & \textbf{10} & \textbf{52} & \textbf{219} & \textbf{-18.00} \\
%ENNL & \textbf{15} & \textbf{61} & \textbf{108} & \textbf{9.84}   & 12          & 68          & 174          & -6.68           \\
%ENZH & \textbf{11} & \textbf{51} & \textbf{187} & \textbf{-10.69} & 10          & 56          & 298          & -30.98          \\
%\hline
%        & \multicolumn{4}{c}{Document}\\
%        \hline
%ENCA        & \textbf{5}  & \textbf{51} & %\textbf{158} & \textbf{-16.99} \\
%\bottomrule
%\end{tabular}
%\caption{Creativity score for two different text granularities: paragraph and document. The best score per language and criterion is shown in bold.}
%\label{tab:table3}
%\end{table}

In this instance, the best solutions for ES and CA are at document level, as we would expect since the context of the sentences is considered. However, for NL and ZH the best performance is at paragraph level, mainly due to error points. 

In the case of NL, the version at document level included more grammatical errors (such as missing articles \textit{van drieling} ("of triplet"), incorrect subject-verb agreement, e.g. \textit{wat je zaken was} ("What your business is", where \textit{zaken} is plural but \textit{was} singular), hallucinations ("formidable" became \textit{ontslagbaar}, a non-existent word, meaning something like \textit{unfireable}), and typos than the version at paragraph level. The paragraph level version does lack consistency at times ("orderly" is translated differently three times), but still has fewer errors.

For ZH, the document-level translations tend to make more grammatical and factual errors, too, especially towards the end of the document. For example, %the 
"sheave-carrier" is translated to "\begin{CJK}{UTF8}{gbsn}
运载屁股的人\end{CJK}
 (ass-carrier)", which is a critical error that disrupts the narrative significantly. "He was seven feet tall" is translated to 
 "\begin{CJK}{UTF8}{gbsn}
 两个高大的人
 \end{CJK} 
 (two tall men)", perhaps in an attempt to convert seven feet to two meters. "He said to her as she fell" is translated to "\begin{CJK}{UTF8}{gbsn}
 他对她说，落
 \end{CJK} 
 (he said to her, falls)" and is not coherent in the target language. Such examples suggest that ChatGPT tends to perform progressively worse for ZH as it processes the whole document. This is in line with previous findings by~\citet{wang-etal-2024-benchmarking} that LLMs demonstrate short-comings in long-text translations, and their performance diminishes as document size increases.

 For CA, the difference (in quantitative terms) between the paragraph and the document levels is largely accounted for by the fact that, in the latter, all the fanciful sobriquets\footnote{These are nicknames given to the gas chambers in this dystopian world, e.g. Weep-no-more, Good-by, Mother or Easy-go} for an institution (the Federal Bureau of Termination) are translated, whereas at paragraph level only 6 (out of 14) are. In other respects, differences between the two CA versions are not that pronounced.
 
 For ES, the paragraph and document level translations are not that dissimilar quantitatively. However, the document level resolves certain translations problems better. For example, %the sentence 
 the expression
 "seven feet tall" is converted at document-level into meters while it remains in feet at sentence level, and %the expression 
 "trick telephone number" is translated as \textit{número de teléfono trampificado} which does not exist as a term, while the document-level uses \textit{número de teléfono con truco} that is correct in Spanish.
 
\subsection{Phase 2. Temperature}\label{s:res_temperature}

Table \ref{tab:table4} shows the results of ChatGPT outputs when the temperature was set %either at 0.0 (2a) or at 1.0 (2b).
at 0.0 (2a) and at 1.0 (2b).

The best performance for ES, NL and ZH are at a temperature of 1.0, but for CA the best output is at temperature 0.0. For most languages, a temperature value of 1.0 outputs more CSs but also more errors--only in ES does a temperature of 0.0 have more errors--as was expected.

In NL, for instance, the output at temperature 1.0 translates "triplets" as \textit{drieën} (\textit{threes})--this is more creative and it could work in some contexts, but not when talking about three babies born at the same time. Still, weighing the CSs against the errors in the creativity index reveals that a temperature of 1.0 has a better output for ES, NL and ZH, despite the errors in the last two.

The general trend is observable for CA too--a higher temperature yields both more CSs and more errors. What sets CA apart %from the other languages 
is that the higher number of CSs does not compensate for the number of errors because of their severity. At temperature 0.0, for example, "Chicago Lying-in Hospital" is adequately translated, whereas at temperature 1.0 the "Lying-in" segment is left untranslated. Other segments are translated in both settings, but the rendering provided at temperature 1.0 is not acceptable. For example, "Kiss this sad world toodle-oo" is translated as \textit{donaré adéu} (‘I will give goodbye'), a collocation that does not exist in CA. %the target language. 
Also, "Good gravy", used as an interjection, is adequately translated at temperature 0.0 and wrongly rendered as \textit{Bona sort} ("Good luck") at %temperature 
1.0.

\begin{table*}[htbp]
    \small
    \centering
    \begin{tabular}{@{}ccccccccc@{}}
    \toprule
        & \multicolumn{4}{c}{T-0.0 (2a)}                             & \multicolumn{4}{c}{T-1.0 (2b)}                                  \\ \midrule
        & \# CSs & \# Errors   & Error Points & Score           & \# CSs      & \# Errors   & Error Points & Score           \\ \midrule
    %ENCA & 3      & \textbf{57} & \textbf{200} & \textbf{-27.67} & \textbf{4}  & 69          & 248          & -33.79          \\
    ENCA & 4      & \textbf{57} & \textbf{200} & \textbf{-25.82} & \textbf{6}  & 69          & 248          & -30.08          \\
    ENES & 8      & 55          & 216          & -21.00          & \textbf{9}  & \textbf{50} & \textbf{164} & \textbf{-11.00} \\
    ENNL & 11     & \textbf{49} & \textbf{99}  & 3.93            & \textbf{12} & 52          & 108          & \textbf{6.13}   \\
    ENZH & 11     & \textbf{45} & \textbf{155} & -5.38           & \textbf{14} & 51          & 165          & \textbf{-1.48}  \\ \bottomrule
    \end{tabular}
    \captionsetup{justification=centering, font=small}
    \caption{Creativity score for two different temperature values: 0.0 and 1.0. \\The best score per language and criterion is shown in bold.}
    \label{tab:table4}
\end{table*}

\begin{table*}[htbp]
    \small
    \centering
    \begin{tabular}{@{}ccccccccc@{}}
    \toprule
        & \multicolumn{4}{c}{Prompt 2 (3a)}                   & \multicolumn{4}{c}{Prompt   3 (3b)}                             \\ \midrule
        & \# CSs     & \# Errors & Error Points & Score  & \# CSs      & \# Errors   & Error Points & Score           \\ \midrule
    %ENCA & \textbf{3} & 57        & 204          & -28.33 & \textbf{3}  & \textbf{55} & \textbf{202} & \textbf{-28.00} \\
    ENCA & \textbf{4} & 57        & 204          & -26.48 & \textbf{4}  & \textbf{55} & \textbf{202} & \textbf{-26.15} \\
    ENES & 12         & 57        & 244          & -18.31 & \textbf{13} & \textbf{43} & \textbf{166} & \textbf{-3.50}  \\
    ENNL & 10         & 43        & 89           & 3.73   & \textbf{18} & \textbf{30} & \textbf{76}  & \textbf{20.71}  \\
    ENZH & 14         & 39        & 171          & -2.48  & \textbf{15} & \textbf{37} & \textbf{161}          & \textbf{1.03}  \\ \bottomrule
    \end{tabular}
    \captionsetup{justification=centering, font=small}
    \caption{Creativity score for two different prompting strategies. \\The best score per language and criterion is shown in bold.}
    \label{tab:table5}
\end{table*}

\subsection{Phase 3. Prompting Strategies}
Table \ref{tab:table5} shows the results for ChatGPT outputs when prompting with more information about genre and author (Prompt 2, 3a) or a request of creativity (Prompt 3, 3b).

For all our languages, Prompt 3 (3b) has better solutions than Prompt 2 (3a) as it generates more CSs and fewer errors. When compared to the results of the other phases, we also see that Prompt 3 has the best performance overall for ES, NL and ZH, with the most number of CSs and the least number of errors. However, for CA, the best performance was in Phase 1 (1b), with Prompt 1 at the document level. % (outputs at paragraph level in Phase 1 and outputs with a temperature of 0.0 in Phase 2 also had higher scores than Prompt 3 for CA). 
The explanation for this lies again in the translation of sobriquets, which are left untranslated in both 3a and 3b. In fact, the only settings in which sobriquets are translated at all are paragraph level (6 out of 14, as said above) and document level (all of them). Since 4 sobriquets are UCPs classified independently, their translations impact the formula. If the sobriquets were excluded, 3a and 3b would be the best-performing settings for CA. The sobriquets were also problematic for ZH and ES: for ES, 3a kept all sobriquets in English and 3b did not translate 2 out of 14 sobriquets; for ZH, it was 3b that did not translate the words but kept them in English, although 3b output had better performance than 3a or any of the other outputs. Surprisingly, for NL, both 3a and 3b  translated the sobriquets into Dutch, although 3a retained one sobriquet in English. This might explain the relative high score for NL with 3b compared to the other languages.  

\subsection{ChatGPT vs Others}\label{s:results_nmt}
We also compare the performance of ChatGPT with that of DeepL and since CA is not available in the latter, we use GT (ENCA-G) and Softcatalà (ENCA-S). At the time we ran DeepL its new-gen version was available for ZH but not for ES nor NL. Therefore we used DeepL new-gen for ZH and DeepL classic for ES and NL. The creativity index of these baseline systems are shown in Table \ref{tab:table2}.

\begin{table}[htbp]
\small
\centering
\begin{tabular}{@{}ccccc@{}}
\toprule
     & \# CSs & \# Errors & Error Points & CI  \\ \midrule
ENCA-S & 1      & 83        & 393          & -63.43 \\
ENCA-G & 1      & 66        & 261          & -41.50 \\
ENES & 9      & 58        & 237          & -22.70 \\
ENNL & 6      & 51        & 103          & -6.00  \\
ENZH & 11     & 42        & 152          & -4.88  \\ \bottomrule
\end{tabular}
\caption{Creativity Index in Others (3c). S stands for Softcatalà's \textit{Traductor} and G for Google Translate.}
\label{tab:table2}
\end{table}

In all languages, the selected NMT system (3c) performs worse than the best setting of ChatGPT. In ZH, NL and ES, DeepL performs better than some of the other settings in ChatGPT, while in CA the two NMT systems perform worse than all ChatGPT outputs. This shows that ChatGPT, with an appropriate prompting strategy, has the potential to outperform its NMT counterparts in literary text in terms of creativity.

\section{Analysis}
\begin{figure}[h]
\centering
\includegraphics[width=0.5\textwidth]{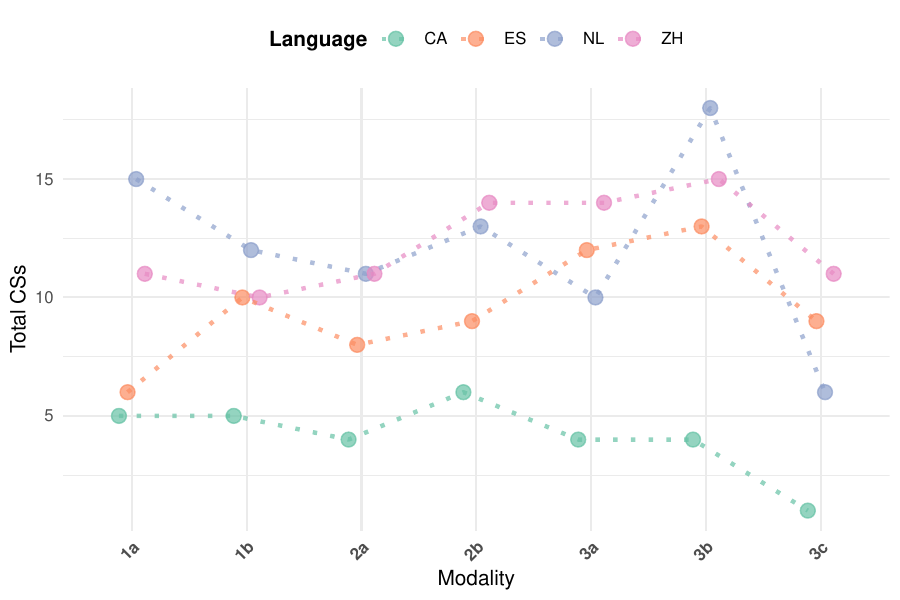}
\caption{Total CSs per Modality and Language}\label{fig:css_gpt}
\end{figure}

\begin{figure}[h]
\centering
\includegraphics[width=0.5\textwidth]{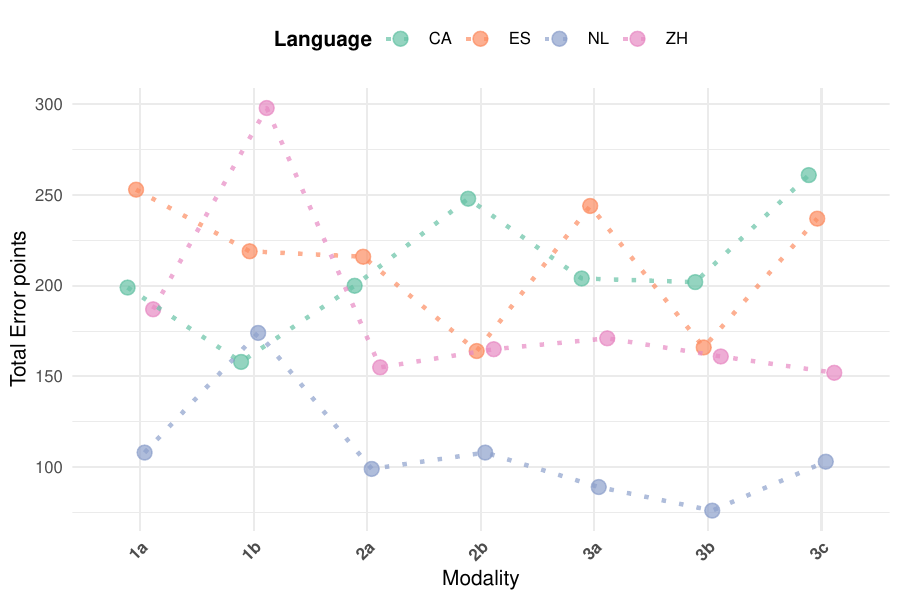}
\caption{Total Error points per Modality and Language}\label{fig:errors_gpt}
\end{figure}

We wanted to further analyse the MT performance in all the phases, and also compare the best performing setting with the professional translation described in Section 3.6. 
Firstly, Figure~\ref{fig:css_gpt} and Figure~\ref{fig:errors_gpt} show the comparison between ChatGPT in the different settings in terms of CSs and Error points (including Others as in Section~\ref{s:results_nmt}).

Figure~\ref{fig:css_gpt} and~\ref{fig:errors_gpt} illustrate the results already illustrated in Tables \ref{tab:table3}, \ref{tab:table4}, \ref{tab:table5}, \ref{tab:table2} more clearly. To assess the effect of Modality and Language on CSs, an Aligned Rank Transform (ART) ANOVA was conducted for non-parametric data. Results show a significant main effect of Language, F(3, 1431) = 30.26, p = .000. However, there are no effects of Modality or the interaction of Modality and Language. Pairwise comparisons using Bonferroni correction show that CSs was significantly lower in CA than ES, NL and ZH p = .000. This is somewhat logical as the number of CSs is very low in all settings, and even lower in CA. We then assess the effect of Modality and Language on Error Points, the results show a significant effect of Modality, F(6, 1431) = 4.63, p = .000, and Modality × Language interaction, F(18, 1431) = 1.7, p = .03. The pairwise comparisons show that Error points was significantly higher in 1b when compared to 2a, p = .000, and to 3b, p = .001, and 2b was significantly higher than 3b, p = .025. This shows again that 1b and 3b were the best performing settings for these languages. The interaction analysis shows only a significant result between 3b/NL and 3c/CA.

\begin{figure}[h]
\centering
\includegraphics[width=0.5\textwidth]{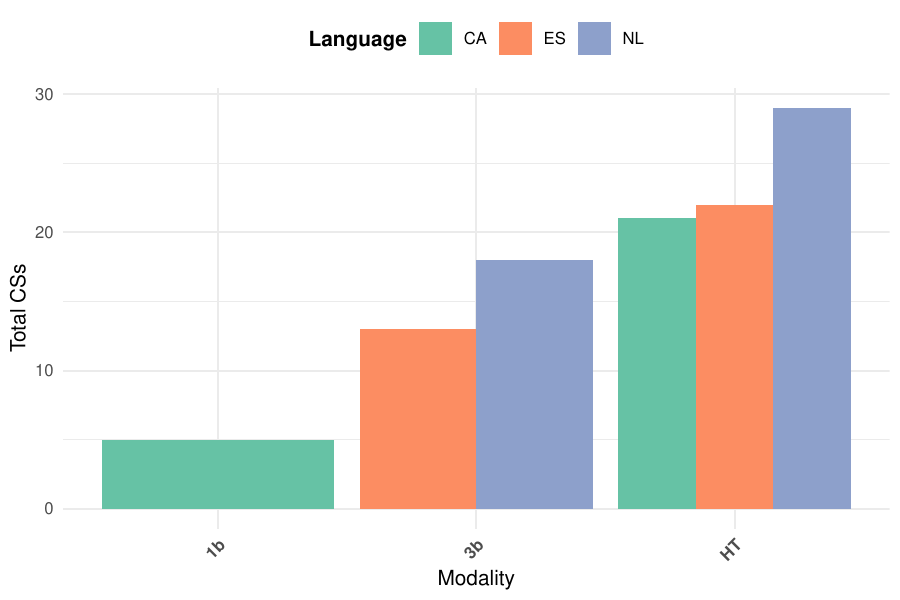}
\caption{Total CSs per best ChatGPT Modality and HT}\label{fig:css_ht}
\end{figure}

\begin{figure}[h]
\centering
\includegraphics[width=0.5\textwidth]{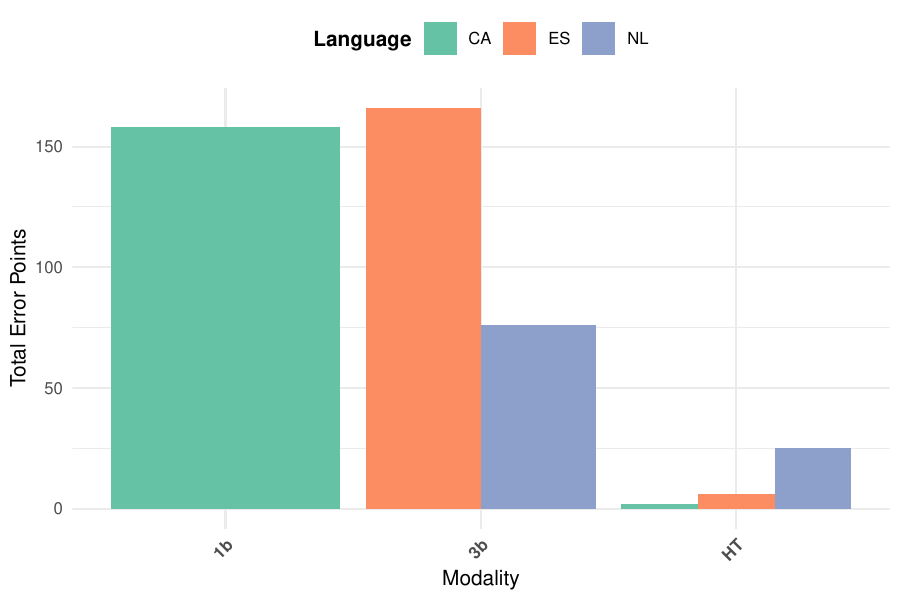}
\caption{Total Error points per best ChatGPT Modality and HT}\label{fig:errors_ht}
\end{figure}

Secondly, %Figure~\ref{fig:css_ht} and Figure~\ref{fig:errors_ht} 
Figures \ref{fig:css_ht} and \ref{fig:errors_ht}
illustrate the comparison of the best performing setting with the professional translations in terms of CSs and Error points. To assess the effect of Modality and Language on CSs, we created a subset by grouping the best performing setting under the variable MT to compare it to HT. The ANOVA indicates a significant main effect of Modality (only HT and MT in this case), F(1, 265) = 31.70, p = .000, and Language, F(2, 265) = 4.26, p = .015. There was no effect of the interaction of Modality and Language. %The 
A
pairwise comparison shows that CS was significantly higher in HT than in MT (p = .00). The effect of Language was only significant for CA and NL (p = .02), but not for CA and ES or ES and NL. 

For Error points the results show a significant main effect of Modality, F(1, 265) = 46.27, p = .00, Language, F(2, 265) = 5.21, p = .006, and Modality × Language interaction, F(2, 265) = 10.66, p = .00. %The 
A
pairwise comparison shows that Error points was significantly lower in HT than in MT (p = .00). The effect of Language was only significant for ES and NL (p = .005), but not for CA and ES or CA and NL. When looking at the interactions, the comparison of HT/Languages and MT/Languages, there is significance in all the combinations of HT and MT (p = .00) except in the interaction between HT/NL and MT/NL.

\subsection{Analysis of AEMs}

Our main interest in running a set of representative AEMs (see Section~\ref{s:data_eval}), is to find out whether any of them correlates significantly with any of the metrics used in the human annotation (i.e. CSs, error points, CI).
A limitation in this regard is that the number of instances% ($n=8$ for CA and $n=7$ for the other three languages) 
\footnote{i.e. number of modalities per language: $n=8$ for CA and $n=7$ for the other three languages.}
is very small, which is why we use a non-parametric correlation metric (Spearman). We find, as expected, significant correlations between pairs of AEMs, which occur most often between pairs of string-based metrics.
Only for one language (NL) do we find significant correlations between one human metric (CSs), and two AEMs: chrF ($p<.001$) and COMET-Kiwi ($p<0.05$).
Given, again, the small sample size, and that they occur only in two cases, we refrain from drawing any strong conclusion.

We also calculated detailed scores for the TER metric.
Namely, the number of operations per operation type (insertions, deletions, shifts and substitutions), system and language.
The main observation is that across all languages and systems, the number of substitutions (range $[960,1286]$) is considerably higher than the number of the other operation types put together: insertions ($[149,290]$), deletions ($[102,225]$) and shifts ($[95,129]$).
All the scores with AEMs are reported in Appendix~\ref{a:aems}.
%\todo{I would include here the table of AEMs metrics and move the correlations to the Appendix B}

\section{Conclusions}
We wanted to explore ChatGPT MT for the best possible setting for creativity. The results show that there is indeed variability per configuration and %also 
per language. 
The first observation, perhaps obvious for a translator but not so obvious for others, is that creativity is seriously affected by using ChatGPT in any setting. Not only is the number of CSs in the TTs provided by all ChatGPT models (but also DeepL, GT and Softcatalà) significantly lower than in HT, but the number of errors is also significantly higher. Even the most creative setting does not come close %to the HT 
%counterparts 
in three out of the four languages analysed %here
(for ZH we did not have an HT reference). %Furthermore, 
Further, it is important to note %here 
that the CI for HT is not only higher but it might also not be representative of the overall creativity of the HT TTs, since we are only analysing the solutions provided by the translators to the annotated UCPs but not the entire segment where translators use other techniques, 
%such a compensation, 
e.g. compensations, to create the desired overall effect of the text.

The second observation is that less appears to be more when prompting ChatGPT to output a creative translation. Overall the best result is the one provided by \textbf{Prompt 3}: ``Translate the following text into [TG] creatively''. Although this prompt still yields a very high number of errors and very modest CSs, it still outperforms the others in ES, NL and ZH while in CA even less information is needed as \textbf{Prompt 1}: ``Translate the following text into [TG]'' outperforms the others. These results are in line with the previous results obtained for Chinese in \citet{du2024thesis}.

The different prompts have somewhat similar results across different languages, with better outputs for temperature 1.0 (2b) and with Prompt 3 (3b) for ES, NL and ZH, although there were differences when providing ChatGPT with paragraphs or the whole document and between CA and the other languages. Moreover, it is interesting to see that there is a level of randomization in the output that is quite unpredictable and that requires many iterations to find the optimal solution. We wonder how this fits in a context where MT is supposed to be used to increase translator\textquotesingle s performance. Trying these different alternatives and still obtaining a sub-optimal result does not seem the best solution for practicing translators, although it is impossible to predict if some MT suggestions might spark creativity.

As this case study is of an exploratory nature, there are limitations, notably, we selected a reduced number of UCPs that were annotated by one single %resource, 
annotator, with a limited number of prompts. However, the striking differences in the performance in literary translation in comparison to what is reported in the media, i.e. singularity \citep{translated_discover_2025}, merits urgent attention.

\section*{Acknowledgments}%Funding}
This project has received funding from the EU ERC Consolidator Grant 101086819; a Beatriz Galindo senior fellowship (BG23/00152) from the Spanish Ministry of Science and Innovation; and Grant PID2023-150711OB-I00 funded by MICIU/AEI/ 10.13039/501100011033 and ERDF/EU.

\section*{Sustainability statement}
All in all, we submitted 2,586 API requests to ChatGPT, leading to the processing of 436,054 tokens (combining inputs and outputs).
To the best of our knowledge, the average CO$^2$ emissions of GhatGPT models is not disclosed. A calculation by a third party estimates that each message sent to ChatGPT produces approximately 4.32g CO$^2$~\citep{co2em}.
Asking ChatGPT we obtain the range $[2.5,23.75]$, depending on the electricity source and assuming 50 Wh per query. Using the 4.32g CO$^2$ figure above, our experiments would have emitted 11.2kg CO$^2$.

It is also worth taking into account that we submit two rather different types of queries: paragraph- and document-based. For a paragraph-based translation we submit 125 queries, which take around 2 minutes and 50 seconds, i.e. 1.36 seconds per query.
For a document-based translation only 1 query is sent, which takes around 2 minutes and 7 seconds.
%ChatGPT
%API requests: 2586 %528 + 1524 + 534
%Tokens: 436,054 %71,504 + 288,926 + 75,624 

% Bibliography entries for the entire Anthology, followed by custom (mtsummit25) entries
%\bibliography{anthology,mtsummit25}
\bibliography{mtsummit25}

\appendix
%\clearpage
%\onecolumn 

\vspace{0.5cm}
\section{Detailed Human Annotations}\label{a:human_annotations}

Tables \ref{t:detailed_enzh}, \ref{t:detailed_ennl}, \ref{t:detailed_enes} and \ref{t:detailed_enca} %6, 7, 8, and 9 
show the detailed human annotations of all languages. The best condition per language is shown in bold.

\begin{table*}[htbp]
    %\small
    %\tiny
    \footnotesize
    \centering
    \scalebox{0.84}{
    \begin{tabular}{@{}cccccccc@{}}
    \toprule
    ENZH           & Para   & Doc          & T-0.0       & T-1.0      & Prompt2     & Prompt3        & DeepL        \\ \midrule
    Abstraction    & 1      & 1            & 1           & \textbf{2} & 0           & 0              & 1          \\
    Concretization & 5      & 5            & 5           & 5          & 4           & \textbf{8}     & 4          \\
    Modification   & 5      & 4            & 5           & 7          & \textbf{10} & 7              & 6          \\
    Reproduction   & 35     & \textbf{30}           & 39 & 35         & 35          & 32             & 36         \\
    Omission       & 4      & 5            & 4           & 2          & 2           & \textbf{1}     & \textbf{1} \\
    Error in UCPs  & 4      & 9            & \textbf{0}  & 3          & 3           & 6              & 6          \\
    \#CSs          & 11     & 10           & 11          & 14         & 14          & \textbf{15}    & 11         \\
    \#Errors       & 51     & 56  & 45          & 51         & 39          & \textbf{37}             & 42         \\
    Error Points   & 187    & 298 & 155         & 165        & 171         & 161            & \textbf{152}        \\
    Score          & -10.69 & -30.98       & -5.38       & -1.48      & -2.48       & \textbf{1.03} & -4.88      \\ \bottomrule
    \end{tabular}
    }
    \caption{\small Detailed human annotation - ENZH}
    \label{t:detailed_enzh}
\end{table*}

\begin{table*}[htbp]
    %\small
    %\tiny
    \footnotesize
    \centering
    \scalebox{0.84}{    
    \begin{tabular}{cccccccc}
    \toprule
    ENNL           & Para        & Doc        & T-0.0 & T-1.0 & Prompt2    & Prompt3        & DeepL        \\ \midrule
    Abstraction    & 2           & \textbf{5} & 3     & 4     & 3          & 2              & 1          \\
    Concretization & \textbf{4}  & 3          & 2     & 2     & 3          & \textbf{4}     & 2          \\
    Modification   & 9           & 4          & 6     & 7     & 4          & \textbf{12}    & 3          \\
    Reproduction   & \textbf{33} & 37         & 40    & 38    & 42         & \textbf{33}    & 44         \\
    Omission       & 0           & 0          & 0     & 0     & 0          & 0              & 1 \\
    Error in UCPs  & \textbf{2}  & 5          & 3     & 3     & \textbf{2} & 3              & 3          \\
    \#CSs          & 15          & 12         & 11    & 13    & 10         & \textbf{18}    & 6          \\
    \#Errors       & 61          & 68         & 49    & 51    & 43         & \textbf{30}    & 51         \\
    Error Points   & 108         & 174        & 99    & 108   & 89         & \textbf{76}    & 103        \\
    Score          & 9.84        & -6.68      & 3.93  & 6.13  & 3.73       & \textbf{20.71} & -6.00      \\ \hline
    \end{tabular}
    }
    \caption{\small Detailed human annotation - ENNL}\label{t:detailed_ennl}
\end{table*}

\begin{table*}[htbp]
    %\small
    %\tiny
    \footnotesize
    \centering
    \scalebox{0.8}{    
    \begin{tabular}{cccccccc}
    \toprule
    ENES           & Para   & Doc        & T-0.0      & T-1.0        & Prompt2    & Prompt3        & DeepL         \\ \midrule
    Abstraction    & 1      & \textbf{2} & 1          & 1            & \textbf{2} & 1              & 1           \\
    Concretization & 2      & 2          & 3          & 2            & \textbf{4} & \textbf{4}     & 1           \\
    Modification   & 3      & 6          & 4          & 6            & 6          & \textbf{8}     & 7           \\
    Reproduction   & 42     & 43         & 44         & 42           & 40         & \textbf{38}    & \textbf{38} \\
    Omission       & 1      & \textbf{0} & 1          & 1            & \textbf{0} & \textbf{0}     & 2           \\
    Error in UCPs  & 5      & \textbf{1} & \textbf{1} & 2            & 2          & 3              & 5           \\
    \#CSs          & 6      & 10         & 8          & 9            & 12         & \textbf{13}    & 9           \\
    \#Errors       & 59     & 52         & 55         & 50           & 57         & \textbf{43}    & 58          \\
    Error Points   & 253    & 219        & 216        & \textbf{164} & 244        & 166            & 237         \\
    Score          & -31.00 & -18.00     & -21.00     & -11.00       & -18.31     &     \textbf{-3.50} & -22.70      \\ \hline
    \end{tabular}
    }
    \caption{\small Detailed human annotation - ENES}\label{t:detailed_enes}

\end{table*}

\begin{table*}[hbtp]
    %\small
    %\tiny
    \footnotesize
    \centering
    \scalebox{0.76}{    
    \begin{tabular}{@{}ccccccccc@{}}
    \toprule
    ENCA           & Para        & Doc             & T-0.0  & T-1.0      & Prompt2 & Prompt3 & Softcatalà         & Google Translate       \\ \midrule
    Abstraction    & 0           & 0               & 0      & 0          & 0       & 0       & 0           & \textbf{1} \\
    Concretization & \textbf{1}  & \textbf{1}      & 0      & \textbf{1} & 0       & 0       & 0           & 0          \\
    Modification   & 4           & 4      & 4      & \textbf{5} & 4          & 4       & 1           & 0          \\
    Reproduction   & 42          & 46              & 47     & 45         & 47      & 47      & \textbf{39} & 45         \\
    Omission       & 1           & 1               & 1      & 1          & 1       & 1       & 1  & 1 \\
    Error in UCPs  & 6           & 2               & 2      & 2           & 2       & 2       & 13          & 7         \\
    \#CSs          & 5           & 5      & 4      &\textbf{6} &4         & 4       & 1           & 1          \\
    \#Errors       & \textbf{51} & \textbf{51}     & 57     & 69         & 57      & 55      & 83          & 66         \\
    Error Points   & 199         & \textbf{158}    & 200    & 248        & 204     & 202     & 393         & 261        \\
    Score          & -23.80      & \textbf{-16.99} & -25.82 & -30.08     & -26.48  & -26.15  & -63.43      & -41.50     \\ \bottomrule
    \end{tabular}
    }
    \caption{\small Detailed human annotation - ENCA}\label{t:detailed_enca}

\end{table*}
%\twocolumn  

%ENCA BEFORE JOSEP'S FINAL VERSION (22012025)
%Abstraction    & 0           & 0               & 0      & 0          & 0       & 0       & 0           & \textbf{1} \\
%Concretization & \textbf{1}  & \textbf{1}      & 0      & \textbf{1} & 0       & 0       & 0           & 0          \\
%Modification   & \textbf{4}  & 3               & 3      & 3          & 3       & 3       & 1           & 0          \\
%Reproduction   & 41          & 43              & 45     & 45         & 45      & 45      & \textbf{35} & 39         \\
%Omission       & 2           & 2               & 2      & 2          & 2       & 2       & \textbf{1}  & \textbf{1} \\
%Error in UCPs  & 6           & 5               & 4      & \textbf{3} & 4       & 4       & 17          & 13         \\
%\#CSs          & \textbf{5}  & 4               & 3      & 4          & 3       & 3       & 1           & 1          \\
%\#Errors       & \textbf{51} & \textbf{51}     & 57     & 69         & 57      & 55      & 83          & 66         \\
%Error Points   & 199         & \textbf{158}    & 200    & 248        & 204     & 202     & 393         & 261        \\
%Score          & -23.80      & \textbf{-18.84} & -27.67 & -33.79     & -28.33  & -28.00  & -63.43      & -41.50     \\ 

\begin{table*}[htbp]%[ht!]
\tiny
\centering
\resizebox{\textwidth}{!}{
\begin{tabular}{lcccccccccccccccc}
\toprule
\textbf{System} & \multicolumn{3}{c}{\textbf{BLEU}} & \multicolumn{3}{c}{\textbf{chrF}} & \multicolumn{3}{c}{\textbf{TER}} & \multicolumn{3}{c}{\textbf{COMET}} & \multicolumn{4}{c}{\textbf{COMET-Kiwi}} \\
\cmidrule(lr){2-4} \cmidrule(lr){5-7} \cmidrule(lr){8-10} \cmidrule(lr){11-13} \cmidrule(lr){14-17}
& ENCA & ENES & ENNL & ENCA & ENES & ENNL & ENCA & ENES & ENNL & ENCA & ENES & ENNL & ENCA & ENES & ENNL & ENZH \\
\midrule
1a & 24.8 & 23.0 & 28.9 & 51.9 & 51.1 & 55.3 & 61.4 & 65.0 & 55.8 & 0.781 & 0.7641 & 0.8254 & 0.7857 & 0.8033 & 0.8223 & 0.8075 \\
1b & 23.1 & 22.3 & 26.3 & 50.4 & 50.3 & 53.2 & 63.0 & 64.7 & 58.1 & 0.7687 & 0.7482 & 0.8101 & 0.7804 & 0.7918 & 0.8037 & 0.6098 \\
2a & 25.6 & 23.2 & 29.5 & 52.0 & 51.0 & 56.3 & 60.6 & 64.5 & 55.2 & 0.7744 & 0.7567 & 0.8281 & 0.7935 & 0.8000 & 0.8252 & 0.8089 \\
2b & 23.4 & 22.0 & 29.0 & 50.8 & 50.4 & 55.9 & 63.0 & 65.2 & 55.7 & 0.7693 & 0.7650 & 0.8248 & 0.7753 & 0.8049 & 0.8252 & 0.8093 \\
3a & 26.0 & 22.5 & 28.6 & 52.3 & 49.9 & 55.5 & 60.6 & 65.8 & 56.8 & 0.7736 & 0.7569 & 0.8253 & 0.7908 & 0.7983 & 0.8276 & 0.8092 \\
3b & 25.4 & 22.7 & 25.0 & 51.7 & 50.2 & 53.8 & 61.3 & 65.2 & 61.6 & 0.7703 & 0.7604 & 0.8238 & 0.7880 & 0.7955 & 0.8163 & 0.8017 \\
3c & 21.7 & 25.1 & 31.8 & 47.9 & 51.7 & 55.9 & 65.4 & 62.5 & 54.3 & 0.7158 & 0.7714 & 0.8257 & 0.7495 & 0.8078 & 0.8301 & 0.8077 \\
3d & 25.3 &       &      & 51.3 &      &      & 61.2 &      &      & 0.7576 &       &       &       &       &       & 0.7714 \\
\bottomrule
\end{tabular}
}
\caption{Scores with a set of AEMs for each system and language pair.}
\label{tab:mt_systems}
\end{table*}

\clearpage
\section{Scores with Automatic Evaluation Metrics}\label{a:aems}

\begin{figure}[H]
\centering
\small
\includegraphics[width=1.0\linewidth]{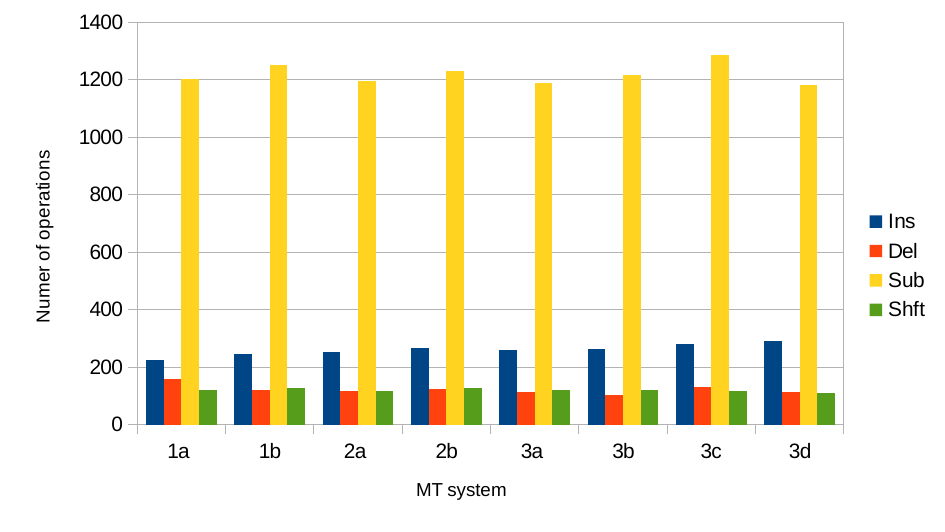}
\caption{TER's number of operations per operation type (insertions, deletions, substitutions and shifts) for English$\Rightarrow$Catalan}\label{fig:ter_enca}
\end{figure}

\begin{figure}[H]
\centering
\small
\includegraphics[width=1.0\linewidth]{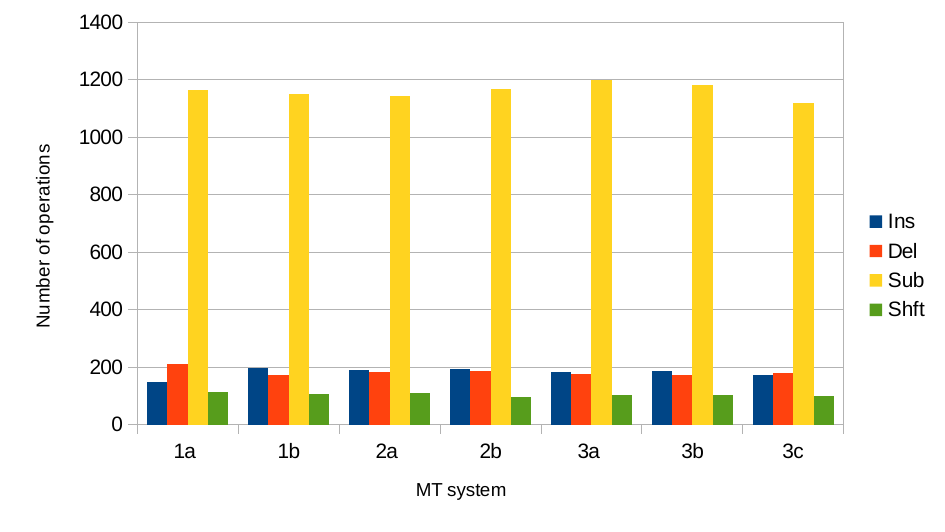}
\caption{TER's number of operations per operation type (insertions, deletions, substitutions and shifts) for English$\Rightarrow$Spanish}\label{fig:ter_enes}
\end{figure}

\begin{figure}[H]
\centering
\small
\includegraphics[width=1.0\linewidth]{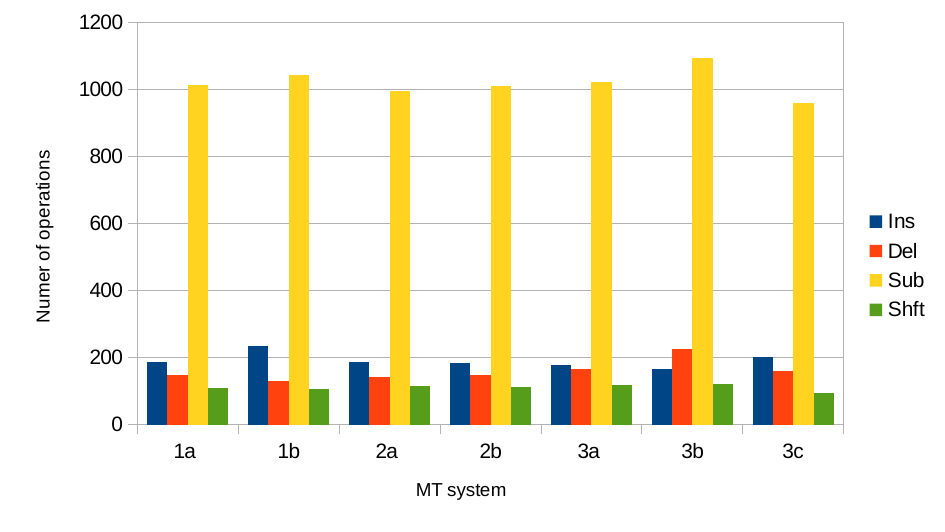}
\caption{TER's number of operations per operation type (insertions, deletions, substitutions and shifts) for English$\Rightarrow$Dutch}\label{fig:ter_ennl}
\end{figure}

Table~\ref{tab:mt_systems} shows the scores for each system and target language with a set of representative automatic evaluation metrics (see Section~\ref{s:data_eval}), while Figure~\ref{fig:ter_enca}, Figure~\ref{fig:ter_enes} and Figure~\ref{fig:ter_ennl}, show TER's number of operations per operation type for each system and target language.

\end{document}